\pdfoutput=1

\documentclass[11pt]{article}

\usepackage[preprint]{coling}

\usepackage{times}
\usepackage{latexsym}
\usepackage{multirow}
\usepackage{threeparttable}
\usepackage{booktabs} 
\usepackage{graphicx}
\usepackage{caption}
\usepackage{subcaption}
\usepackage{svg}
\usepackage{tabularx}
\usepackage{tikz}
\usepackage{graphicx}
\usetikzlibrary{shapes.geometric, arrows.meta, positioning}
\usepackage{amsmath} 
\usepackage[T1]{fontenc}

\usepackage[utf8]{inputenc}

\usepackage{microtype}

\usepackage{inconsolata}

\usepackage{graphicx}

\usepackage{xspace}
\newcommand{\dsltl}{{\textsc{ DSL-TL}}\xspace}
\newcommand{\cupansp}{{\textsc{ CubanSpVariety}}\xspace}

\title{Common Ground, Diverse Roots: The Difficulty of Classifying Common Examples in Spanish Varieties}

\author{Javier A. Lopetegui\(^{*}\) \quad Arij Riabi\(^{*}\) \quad Djamé Seddah\\
	INRIA Paris, France \\
	\texttt{firstname.lastname@inria.fr}
}

\begin{document}
\maketitle

\renewcommand\thefootnote{\fnsymbol{footnote}}
\footnotetext[1]{These authors contributed equally.}
\renewcommand*\thefootnote{\arabic{footnote}}
\begin{abstract}

Variations in languages across geographic regions or cultures are crucial to address to avoid biases in NLP systems designed for culturally sensitive tasks, such as hate speech detection or dialog with conversational agents. In languages such as Spanish, where varieties can significantly overlap, many examples can be valid across them, which we refer to as common examples. Ignoring these examples may cause misclassifications, reducing model accuracy and fairness. Therefore, accounting for these common examples is essential to improve the robustness and representativeness of NLP systems trained on such data.
In this work, we address this problem in the context of Spanish varieties. We use training dynamics to automatically detect common examples or errors in existing Spanish datasets. We demonstrate the efficacy of using predicted label confidence for our Datamaps \cite{swayamdipta-etal-2020-dataset} implementation for the identification of hard-to-classify examples, especially common examples, enhancing model performance in variety identification tasks. Additionally, we introduce a Cuban Spanish Variety Identification dataset with common examples annotations developed to facilitate more accurate detection of Cuban and Caribbean Spanish varieties. To our knowledge, this is the first dataset focused on identifying the Cuban, or any other Caribbean, Spanish variety.

\end{abstract}

\section{Introduction}
Language reflects culture and identity, while also capturing subtle variations that shape communication. In Natural Language Processing (NLP), it is crucial to account for these nuances, especially in language variety identification, where small shifts in meaning, often tied to cultural interpretations, can impact sensitive tasks like hate speech detection. Expressions that may be benign in one dialect can be offensive in another, making accurate variety identification essential to prevent misclassifications and ensure culturally appropriate responses \citep{nozza-2021-exposing, hershcovich-etal-2022-challenges}.
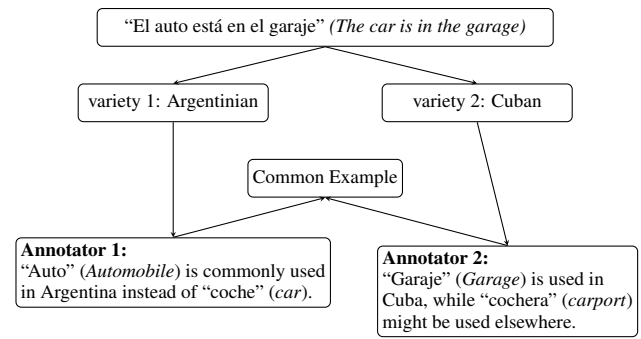
\begin{figure}[ht!]
\centering
\scalebox{0.5}{
\begin{tikzpicture}[
  roundnode/.style={rectangle, draw=black, thick, minimum size=10mm, text centered, rounded corners},
  arrow/.style={thick,->,>=stealth},font=\Large
]

\node[roundnode, minimum width=12cm] (sentence) at (0,5) {``El auto está en el garaje'' \textit{(The car is in the garage)}};

\node[roundnode, minimum width=5cm] (variety1) at (-4,3) {variety 1: Argentinian};
\node[roundnode, minimum width=5cm] (variety2) at (4,3) {variety 2: Cuban};
\node[roundnode, minimum width=4cm] (common) at (0,1) {Common Example};

\draw[arrow] (sentence.south) -- (variety1.north);
\draw[arrow] (sentence.south) -- (variety2.north);

\node[roundnode, minimum width=6cm, align=left] (explain1) at (-4,-1.5) {\textbf{Annotator 1:} \\ ``Auto'' (\textit{Automobile}) is commonly used\\ in Argentina instead of ``coche'' (\textit{car}).};
\node[roundnode, minimum width=6cm, align=left] (explain2) at (4.8,-2) {\textbf{Annotator 2:} \\ ``Garaje'' (\textit{Garage}) is used in\\ Cuba, while ``cochera'' (\textit{carport})\\ might be used elsewhere.};

\draw[arrow] (variety1.south) -- (explain1.north);
\draw[arrow] (variety2.south) -- (explain2.north);
\draw[arrow] (explain2.north) -- (common.south);
\draw[arrow] (explain1.north) -- (common.south);
\end{tikzpicture}}
\caption{Common Example Identification in Language Variety Classification}
\label{fig:example_classification}
\end{figure}
In such tasks, cross-lingual models often struggle with these subtle cultural and linguistic distinctions, as the same formulation may carry vastly different meanings across varieties. Language-specific models tend to perform better in such cases, as they are more sensitive to regional variations \citep{nozza-2021-exposing, vaidya-etal-2024-clteam1, arango2021cross-lingual, montariol-etal-2022-multilingual, castillo-lopez-etal-2023-analyzing}. However, distinguishing between closely related languages, dialects, and regional varieties of the same language is a key and difficult task in language identification \cite{tiedemann-ljubesic-2012-efficient,lui-cook-2013-classifying,Zampieri_Nakov_2021,espana-bonet-barron-cedeno-2024-elote}. Adding to this complexity is the issue of common examples —valid phrases across multiple dialects or varieties. Overlooking these examples can result in biased classifications, especially in languages like Spanish, where variety overlap is frequent. \footnote{Following \citet{Hudson_1996}, we use the terms varieties of Spanish: ``a variety is a set of linguistic items with similar social (including geographical and cultural) distribution.''} Despite this, many current datasets treat the identification of the language variety as a single label classification task, which overlooks this crucial aspect \citep{zampieri-etal-2024-language}.
Current datasets for language variety identification often rely on manual annotations or automated methods such as geographic information \cite{zampieri-etal-2019-report,abdul-mageed-etal-2020-nadi,abdul-mageed-etal-2022-nadi,aepli-etal-2022-findings} or keyword-based classification \cite{Althobaiti_2022}. However, both approaches have limitations, and manually checking large datasets for common examples is challenging and costly \citep{keleg-magdy-2023-arabic, bernier-colborne-etal-2023-dialect}. Datamaps based on training dynamics \citep{swayamdipta-etal-2020-dataset,weber-genzel-etal-2024-varierr}, which track how the confidence of the model changes over epochs, have been used successfully to detect annotation errors and human label variation. These methods highlight which examples are consistently easy or difficult for the model, with hard examples often pointing to ambiguity or errors.
We propose using training dynamics to detect common examples in language variety identification tasks. In \ref{fig:example_classification} we show an example of these common examples.  These are expected to be among the hard examples the model struggles with during training. By tracking the model's confidence in its predicted labels over multiple training epochs, rather than using gold labels, we aim to detect ambiguous instances that are hard for the model to classify consistently.
Our research addresses the following questions:
\begin{itemize}
    \item \textbf{RQ1:} Can training dynamics help detect common examples that are hard for the model to classify during the training?
    \item \textbf{RQ2:} Can we use the model's confidence over predicted labels to detect common examples?
    \item \textbf{RQ3:} Can this approach work effectively across different domains, such as news articles and user-generated content?
\end{itemize}

To investigate these questions, we use two datasets: the Spanish subset of \dsltl dataset \citep{zampieri-etal-2024-language}, which contains texts extracted from news articles, and a new dataset of Cuban Spanish varieties we collected from Twitter. We adapt the Datamaps technique by changing the way confidence and variability are calculated, allowing us to identify common examples. Our results demonstrate the efficiency of this approach in detecting common examples in both datasets.

Our main contributions are as follows:
\begin{enumerate}

    \item We propose a modified Datamaps model that calculates confidence and variability based on the predicted label's probability, providing a more accurate reflection of model uncertainty. Our model can help accelerate the re-annotation of existing datasets.
    \item Using both frequency-based methods and SHAP analysis \citep{lundberg2017unified}, we provide a thorough error analysis that demonstrates the usefulness of our approach to capture annotation errors and shows how the model predictions are topic-dependent. 
    \item We present and publicly share a novel Cuban Spanish variety identification dataset, consisting of 1,762 manually annotated tweets by three native speakers, with labels assigned based on agreement and covering Cuban, non-Cuban varieties, and common examples.
\end{enumerate}

\section{Related Work}

\textbf{Common Examples.} The challenge of handling common examples that can be valid across multiple language varieties has been a recurring issue in language variety identification. Traditional single-label classification often struggles to assign unique labels to common examples\cite{Althobaiti2020AutomaticAD,bernier-colborne-etal-2023-dialect}. Addressing this challenge, \citet{zampieri-etal-2024-language} introduced a third class specifically for common instances in their \dsltl dataset for language variety identification. This dataset allowed the exploration of the impact of these ambiguous cases on model performance. The authors found that the models had difficulty distinguishing between common and dialect-specific examples. Then, their results served as a baseline for the DSL-TL shared task at VarDial 2023 \citep{aepli-etal-2023-findings}. In the scope of this shared task, \citet{vaidya-kane-2023-two} introduced a two-stage multilingual dialect detection system. Their approach first identifies the macro-language, followed by applying dialect-specific models to refine the classification. Although this system performed well overall, it struggled with the common examples class, where it frequently misclassified examples due to the lack of clear dialect-specific markers. 
The Spanish language, with its rich array of varieties, provides a particularly challenging landscape for variety identification due to the high similarity between varieties. \citet{zampieri-etal-2024-language} noted that the prevalence of common examples in Spanish is especially high. Given the significant lexical and syntactical overlap among Spanish varieties, sentences that can belong to more than one variety are frequent, making traditional classification approaches less reliable. The misclassification of these common instances not only introduces noise into the datasets but also impacts the overall performance of the models, as evidenced by the poor handling of Argentine examples in \citet{vaidya-kane-2023-two}.

\paragraph{Multi-class Approaches for Variety Identification.} In light of these challenges that affect many different languages, several works have proposed moving away from single-label classification towards multi-class or multi-label approaches for variety identification. For example, \citet{keleg-magdy-2023-arabic} demonstrated that many sentences could validly belong to multiple Arabic dialects, arguing for including multiple labels per instance. They introduced the Expected Maximal Accuracy (EMA) metric to measure the upper-bound accuracy in scenarios where common instances occur frequently. Their results indicated that the majority of false positives in traditional single-label classifiers were, in fact, not errors, but cases where multiple dialects could be correct.
\citet{bernier-colborne-etal-2023-dialect} took this further by employing similarity metrics to identify duplicate or nearly duplicate examples and assigning multiple labels to ambiguous sentences. Their work, focusing on French varieties, showed that this multi-class approach significantly improved F1-macro scores for ambiguous examples. They argued that applying a multi-class framework can improve the accuracy of variety identification and better handle the inherent ambiguity found in multilingual datasets.

\section{Task Definition: Automatic Common Examples Detection}

Our main task is to identify common examples across similar language varieties. Our proposed pipeline can be separated into two main stages:

\begin{itemize}
    \item Fine-tune a Transformer-based model on the Variety Identification datasets for single-label classification of varieties (binary).
    \item Assign a score to each example using a scorer model, expecting higher values for common examples, and rank them with the highest scores at the top.
\end{itemize}
\subsection{Scorer Models}
\paragraph{Datamaps} \citet{swayamdipta-etal-2020-dataset} proposed Datamaps (DM) using Training Dynamics, which is the behavior of a model as training progresses, for detecting annotation errors in datasets. Their approach focused on tracking the confidence and variability on the gold label during training. Specifically, examples consistently showing low confidence for this label across epochs were flagged as potential annotation errors or ambiguous cases. This technique has also been adapted to identify the variation of human labels, where examples can legitimately belong to more than one category \citep{weber-genzel-etal-2024-varierr}. We use this technique to identify common examples for the Variety Identification task.

\paragraph{Datamaps using predicted label probability} We adapt the Datamaps metrics to our use case. Unlike \citet{swayamdipta-etal-2020-dataset}, who focus on the gold labels, and \citet{weber-genzel-etal-2024-varierr}, who prioritize re-annotating erroneous labels, our goal is to detect instances that the model struggles to classify consistently. Therefore, we calculate confidence and variability differently: rather than focusing on the correctness of assigned labels or identifying annotation errors, we calculate these metrics based on the maximum predicted probability for each instance at each epoch, aiming to detect instances that exhibit inconsistent predictions or low confidence and, therefore, could belong to both classes or an unobserved third class. For common examples, which can be associated with more than one label, it would be more natural to describe the uncertainty in terms of the model's confidence in its predictions.
The \textit{confidence} is defined as:
    \begin{equation}
        \text{DM}_{\text{mean}-\text{pred}}\ =\ - \frac{1}{E}\sum_{e=1}^E \max_{j}(p_{i,j,e})
    \end{equation}
where $p_{i,j,e}$ is the probability assigned to the $i$'th instance for the label $j$ in epoch $e$. Then, the lowest confidences correspond to a higher score because of the negative sign. The idea is that examples with small probabilities associated with the predicted label across the epochs are likely challenging examples. 

The \textit{variability} is defined as:
    \begin{equation}
    \begin{split}
                    \text{DM}_{\text{std}-\text{pred}} =\\ \sqrt{\frac{1}{E}\left(\sum_{e=1}^E{\max_{j}(p_{i,j,e})+ \text{DM}_{\text{mean}-\text{pred}}}\right)^2}\end{split}
    \end{equation}

The high \textit{variability} indicates that the model's confidence changes significantly across epochs, suggesting the model is uncertain about the instance. This can point to an instance that is hard to classify or potentially common.
\paragraph{Random baseline} We use a random model as a scorer, which assigns uniformly random scores between 0 and 1 to each example as a baseline. 

\paragraph{Language Model} For the Variety Identification module we use the model  BETO, a monolingual Spanish BERT model version \citep{canete2020spanish} for our experiments; it has proven effective in several downstream tasks for this language. This model was trained on all Wikipedia and all Spanish data from the OPUS project \cite{tiedemann-2012-parallel}. In the case of Spanish Wikipedia, by 2017, around 39.2\% of edits came from Spain \citep{wiki:spanishwiki}, which can negatively impact the model performance in varieties not from Spain.
\paragraph{Evaluation}

The first metric considered for evaluation is the Average Precision Score in the Common Examples Identification Task. In addition, we evaluate precision and recall by considering the top N instances, ranked by their score values, with N ranging from 10 to the size of each dataset.

\section{Datasets}\label{sec:datasets}
In this section, we describe the datasets used for our analysis. We use an existing dataset \dsltl and propose a new dataset \cupansp focused on the Cuban Spanish variety.

\subsection{\dsltl}
The Discriminating Similar Language - True Labels (\dsltl) dataset \citep{zampieri-etal-2024-language} was employed in a shared task at the VarDial 2023 workshop\footnote{ \href{https://sites.google.com/view/vardial-2023/shared-tasks}{VarDial 2023 website.}}. This dataset contains examples from Portuguese, Spanish, and English varieties, but our focus is solely on the Spanish subset. The Spanish subset is derived from the DSLCC dataset \citep{tan2014merging} and includes sentences extracted from various Argentinian and Spanish newspapers, with each example annotated based on the country associated with the news source. However, annotating the examples with a single label proved difficult, even for human annotators \citep{goutte-etal-2016-discriminating}. Specifically, Spanish annotators achieved an average accuracy of only 54.90\%. To address these challenges, \citet{zampieri-etal-2024-language} randomly sampled the Spanish, Portuguese, and English subsets and conducted a new round of human annotations. In addition to the original binary labels, a third label—\textit{both or neither}—was introduced. This additional label was assigned when annotators were unable to identify the characteristics of the different varieties. For our experiments with the \dsltl dataset, we use the newly introduced labels from the \dsltl dataset and the original labels from the DSL-2014 corpus. It allowed us to simulate a scenario where new annotations would be unavailable. We only use the training set to analyze the training dynamics.

\begin{figure}[hb!]
    \centering
    \begin{subfigure}{\columnwidth}
        \centering        \includegraphics[width=\columnwidth]{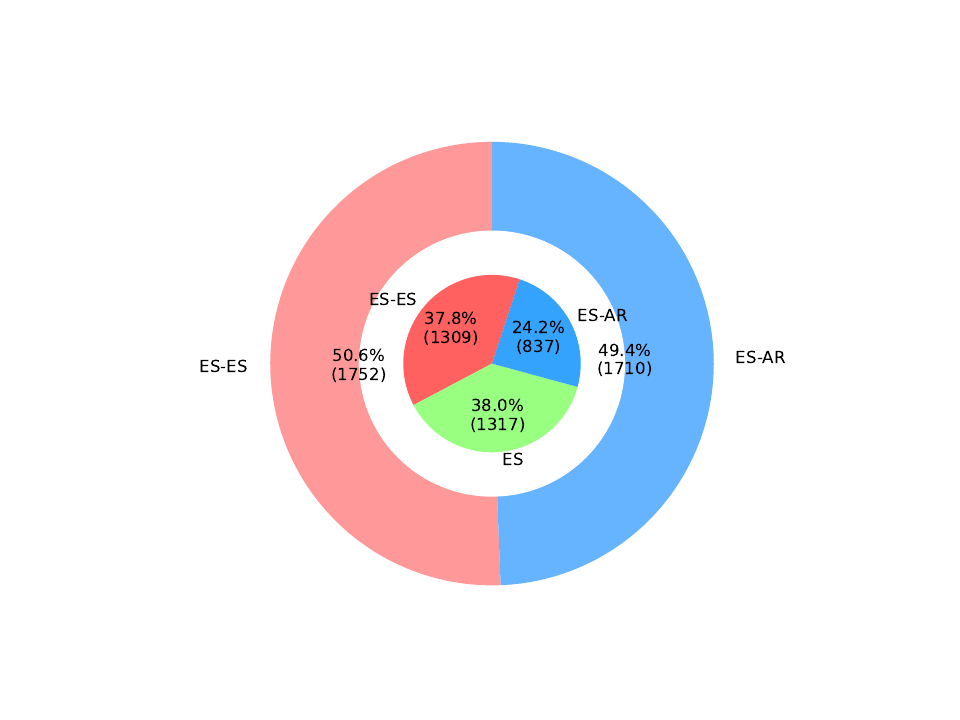}
        \caption{\dsltl dataset distribution.}
        \label{fig:dsl-tl_distribution}
    \end{subfigure}
    \begin{subfigure}{\columnwidth}
        \centering
        \includegraphics[width=\columnwidth]{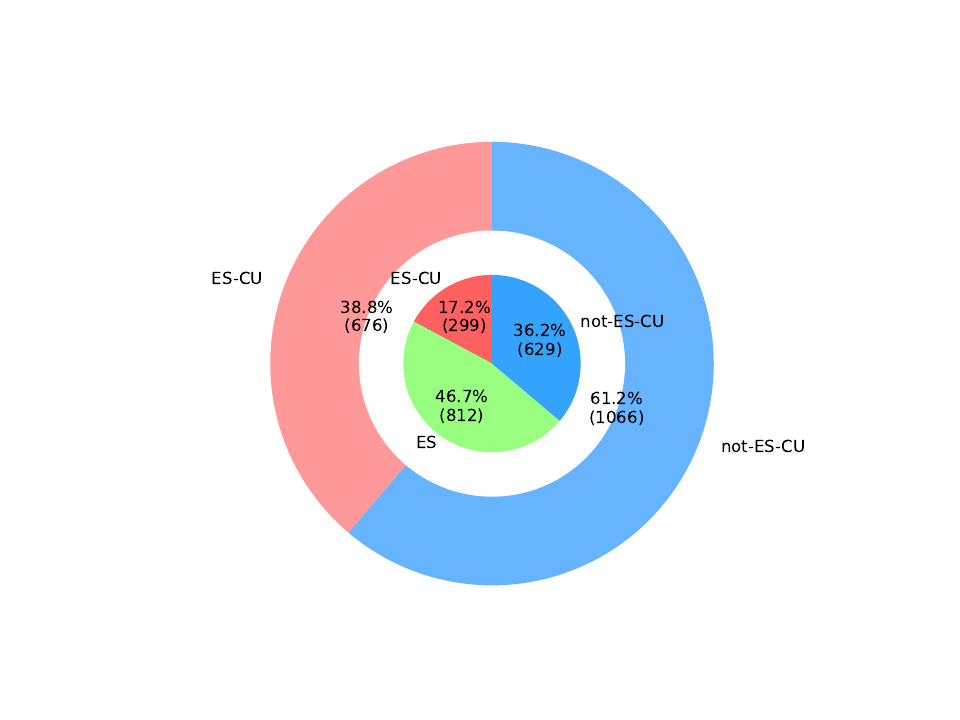}
        \caption{\cupansp dataset distribution.}
        \label{fig:cuban_dataset_distribution}
    \end{subfigure}
    \caption{Datasets distributions.}
    \label{fig:datasets_distribution}
\end{figure}

\subsection{\cupansp}\label{subsec:cuban_variety_dataset}

To our knowledge, the dataset is the first dataset for Cuban or any Caribbean Spanish variety identification. The dataset contains manually annotated tweets with variety information. We collected the data from the publicly available archive \textit{The Twitter Stream Grab} in the website archive.org. We worked particularly with data from July 2021. \footnote{ \href{https://archive.org/details/archiveteam-twitter-stream-2021-07}{Link to available data for July 2021.}}

\paragraph{Data Annotation.} We randomly sampled $10000$ tweets from July 11th and July 12th. Among those, we finally annotated 1762 examples. We considered this time frame because of the high Twitter activity in Cuba after July 11th protest in 2021 with trending hashtags such as \textit{\#SOSCuba} or \textit{\#SOSMatanzas}. \footnote{ \href{https://www.nytimes.com/2021/07/11/world/americas/cuba-crisis-protests.html}{New York Times (July 11th, 2O21).}} Each tweet was annotated across five columns: \textit{cuban\_variety}, \textit{not\_cuban\_variety}, \textit{specific\_variety}, \textit{not\_able\_to\_identify}, and \textit{irrelevant}. Annotators marked \textit{cuban\_variety} if the tweet belonged to the Cuban Spanish variety and \textit{not\_cuban\_variety} if it did not (cf. Section \ref{sec:annotation_guidelines}). In case of identifying a different Spanish variety (e.g., from Spain or Chile), they were asked to annotate it in the \textit{specific\_variety} column for future work. When uncertain about the variety, they marked \textit{not\_able\_to\_identify}. Tweets deemed noisy or non-Spanish were marked as \textit{irrelevant}.

We focused on three labels for analysis: \textit{ES-CU} (Cuban variety), \textit{not-ES-CU} (non-Cuban), and \textit{ES} (common examples). Tweets with \textit{cuban\_variety} marked True were labeled \textit{ES-CU}, those with \textit{not\_cuban\_variety} marked True were labeled \textit{not-ES-CU}, and tweets marked only as \textit{not\_able\_to\_identify} were labeled \textit{ES}, aligning with the \dsltl dataset. Three volunteers, native Cuban Spanish speakers with a Master's degree in Cuba, performed the annotations. Their familiarity with other Spanish varieties helped them recognize common examples. Labels were assigned when at least two annotators agreed and tweets marked as irrelevant by any annotator were discarded.  Full agreement was reached for 984 examples (55.8\%), partial agreement for 776 (43.5\%), with disagreement in just 12 cases (0.7\%).
We use the full dataset for training dynamics analysis. In this case, we only have the annotations with the common examples information (i.e. not single label approach). Then, to simulate a real-world scenario with single labels, we randomly assigned each common example a label of either \textit{ES-CU} or \textit{not-ES-CU}. Figure \ref{fig:cuban_dataset_distribution} shows the final dataset distribution. The internal circle represents the original distribution (cf. Table~\ref{tab:dataset_overview} for an overview of lexical properties).

\section{Results}

\begin{table*}[htb!]
\centering
\footnotesize
\begin{tabular}{cccccc}
\toprule

\multicolumn{1}{r}{Model} & APS & Prec-500 & Recall-500 & Prec-1000 & Recall-1000 \\ \midrule
\multicolumn{6}{c}{\dsltl}\\
Random                              & 39.45 ± 2.54  &  38.71 ± 1.49    & 14.98 ± 0.57              & 37.80 ± 1.16           & 28.99 ± 0.89\\ 
$DM_{mean-pred}$                               & \textbf{54.75} \textbf{±} \textbf{1.8}  & \textbf{62.78 ± 2.47} & \textbf{24.31 ± 0.95} & \textbf{57.76 ± 1.58} & \textbf{44.29 ± 1.21} \\
$DM_{std-pred}$                               & $52.88 \pm 3.00$ & 58.70 ± 3.05 & 22.73 ± 1.18 & 56.03 ± 2.59 & 42.97 ± 1.98 \\
\midrule
\multicolumn{6}{c}{\textbf{\cupansp}}\\
Random                              & 46.42 ± 1.20  & 46.39 ± 2.32 & 29.10 ± 1.46             & 46.83 ± 0.52          & 58.17 ± 0.65\\ 
$DM_{mean-pred}$                               & \textbf{63.51 ± 2.56}  & \textbf{66.19 ± 3.43} & \textbf{41.52 ± 2.15} & \textbf{59.16 ± 1.25} & \textbf{73.50 ± 1.55} \\
$DM_{std-pred}$                               & 61.97 ± 2.60 & 64.86 ± 3.59 & 40.68 ± 2.25 & 58.15 ± 1.07 & 72.25 ± 1.33 \\
\bottomrule
\end{tabular}
\caption{Evaluation metrics for Automatic Common Examples on \dsltl and \cupansp datasets. We present the Average Precision Score, equivalent to the area under the precision-recall curve, and the precision and recall for Top-500 and Top-1000 instances. All the metrics are expressed in percentages.}
\label{table:prob_output}
\end{table*}
\subsection{Variety Identification}
\begin{figure}[htb!]
    \centering
    \includegraphics[width=\linewidth]{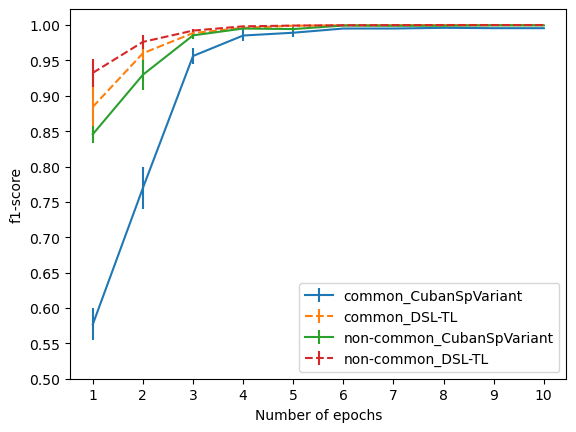}
    \caption{F1-score during training for common and non-common examples on both datasets.}
    \label{figure:train-metrics_all_datasets}
\end{figure}
We investigate the learning behavior of BETO-based Variety Identification model by analyzing the F1 scores across both datasets. Figure \ref{figure:train-metrics_all_datasets} presents the F1-score evaluation for Language Variety Classification over 10 training epochs, with separate curves for common examples and the rest of the data in both datasets. As shown in the figure, the performance gap between common and non-common examples is substantial during the early stages of training. Furthermore, the error bars indicate greater variability in the F1-scores for common examples than the rest. This gap is particularly pronounced in the \cupansp dataset, which exhibits lower F1 scores, likely due to the additional challenges of social media content, unlike \dsltl, which contains sentences from newspaper articles.
These observations suggest that the model finds it more challenging to learn common examples, supporting the idea that their characteristics can be identified through training dynamics.

\subsection{Common Examples Identification}\label{sec:common_resutls}
\begin{figure}[htb!]

         \centering
    \includegraphics[width=\columnwidth]{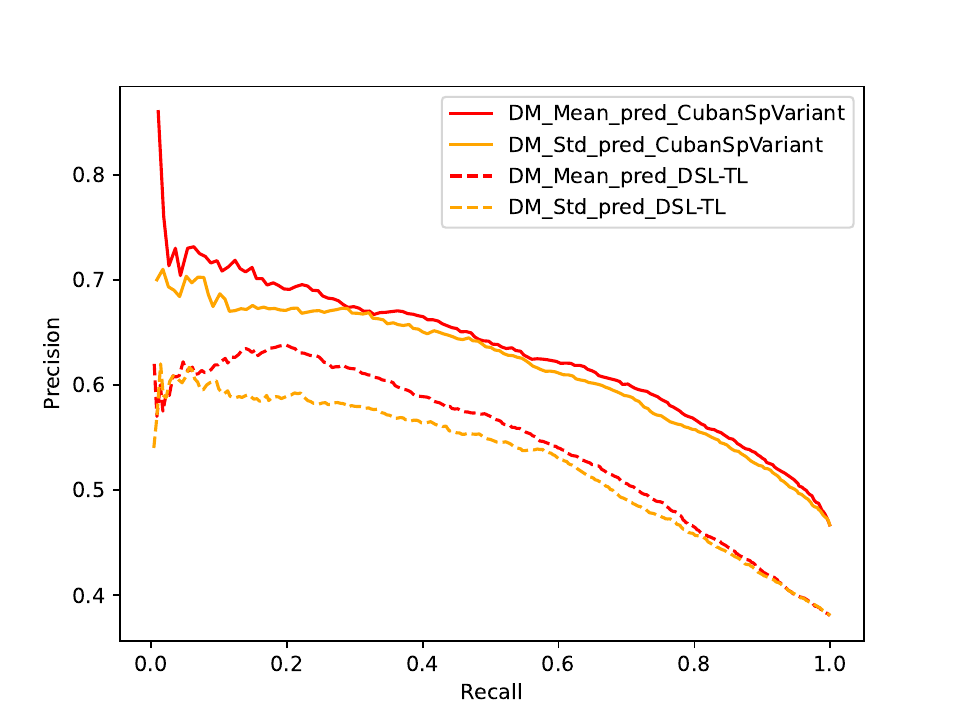}
         \caption{Precision versus recall curve}
         \label{fig:prec_vs_recall}
    \end{figure}
We present in Table \ref{table:prob_output} the results for both the \dsltl and \cupansp datasets, comparing $DM_{\text{mean-pred}}$, $DM_{\text{std-pred}}$ and the random baseline. Across both datasets, the two Datamaps models significantly outperform the baseline, indicating that both capture relevant information about common examples. In addition, $DM_{\text{mean-pred}}$, which leverages the confidence in predicted labels, consistently outperforms $DM_{\text{std-pred}}$. This suggests that the model’s average confidence offers a more reliable signal for identifying common examples, while the variability-based approach ($DM_{\text{std-pred}}$) tracks changes that do not always correspond with common examples.
We observe that the difference in performance between the two datasets follows a similar pattern across all models, including the random baselines. This is likely due to the proportion of common examples in each dataset. In \dsltl, where 38\% of the examples are common, the random baseline precision is close to 38\%. Similarly, in \cupansp, with 46\% common examples, the baseline precision is near 46\%. This suggests that the metrics' ranges are closely tied to each dataset's distribution of common examples. 

Figure \ref{fig:prec_vs_recall} shows both datasets' precision versus recall curves. In both cases, precision remains relatively stable in the early ranking stages and begins to converge toward the common examples' proportion as recall increases. The performance difference between $DM_{\text{mean-pred}}$ and $DM_{\text{std-pred}}$ is more pronounced for smaller values of N, particularly in precision. However, the recall curves show a steeper slope at earlier ranking stages, which gradually decreases as N increases, consistent with expected behavior.
 
    \begin{figure}[htb!]
    \includegraphics[width=\columnwidth]{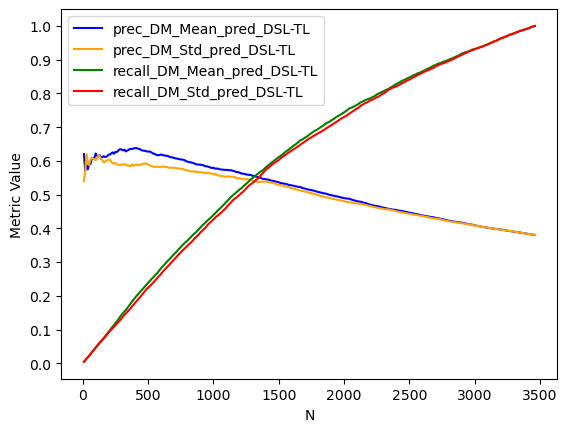}
         \caption{Precision and Recall versus Top-N instances \dsltl dataset}
         \label{fig:precision_recall_vs_N_dsl}
    \end{figure}
  \begin{figure}
         \centering  \includegraphics[width=\columnwidth]{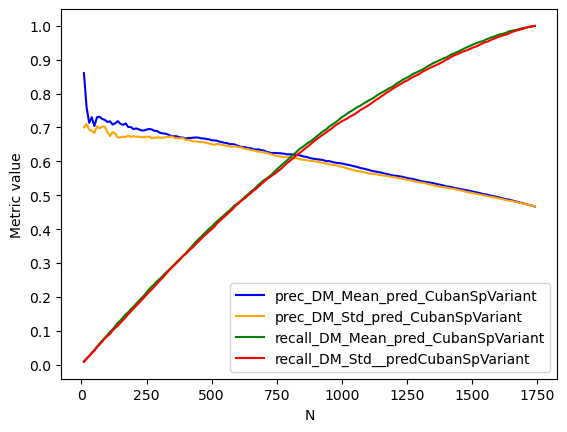}
         \caption{Precision and Recall versus Top-N instances \cupansp dataset}
         \label{fig:precision_recall_vs_N_Cuban}
\end{figure}
    
Figure \ref{fig:precision_recall_vs_N_dsl} highlights that in the \dsltl dataset, which contains clean, edited content unlike our Twitter-based Cuban dataset, $DM_{\text{mean-pred}}$ identifies common examples early in the ranking. This is likely because we had access to the original labels for common examples in this dataset, reducing noise. Furthermore, the clear class boundaries distinguishing Spanish varieties from Spain and Argentina likely contributed to the model's more stable performance, while $DM_{\text{std-pred}}$ is less effective in this context.
In Figure \ref{fig:precision_recall_vs_N_Cuban}, we observe that for the \cupansp dataset, which contains more dynamic and informal language from user-generated content, the performance gap between $DM_{\text{mean-pred}}$ and $DM_{\text{std-pred}}$ becomes smaller. This indicates that variability has a more significant impact on identifying common examples in user-generated content. In this dataset, common examples were identified in the first round and randomly assigned to Cuban or non-Cuban classes, increasing ambiguity.
It is worth noting that, beyond the differences in the nature of the dataset (newswire text vs. Twitter user-generated content), the collection period dates vary over six years between both datasets, likely affecting model performance since languages evolve and are shaped by social dynamics. Furthermore, the Cuban dataset includes tweets from July 11th and 12th, during large protests in Cuba that were trending among Spanish-speaking countries. This may introduce biases into the dataset and influence the variety identification.
\section{Error Analysis}\label{sec:errors}

To better understand our models' performance, we analyzed the errors for each dataset by counting the most frequent words in the Top-500 non-common instances predicted by the $DM_{mean-pred}$ model (prediction errors). After removing stopwords and special tokens, we found that in the \cupansp dataset, the most frequent words were \textit{Cuba} and \textit{SOSCuba}, directly tied to the Cuban variety in this context. In contrast, the \dsltl dataset showed common words like \textit{ha}, \textit{pero}, \textit{fue}, and \textit{también}, which do not indicate a specific variety. The topic bias in the Cuban dataset can influence the model predictions, mainly when the examples contain keywords specific to the variety. This also explains why $DM_{std-pred}$ performs better for \cupansp, as these keywords in both classes make variability more significant than in \dsltl.

\begin{figure}[htb!]
    \centering
    \includegraphics[width=\columnwidth]{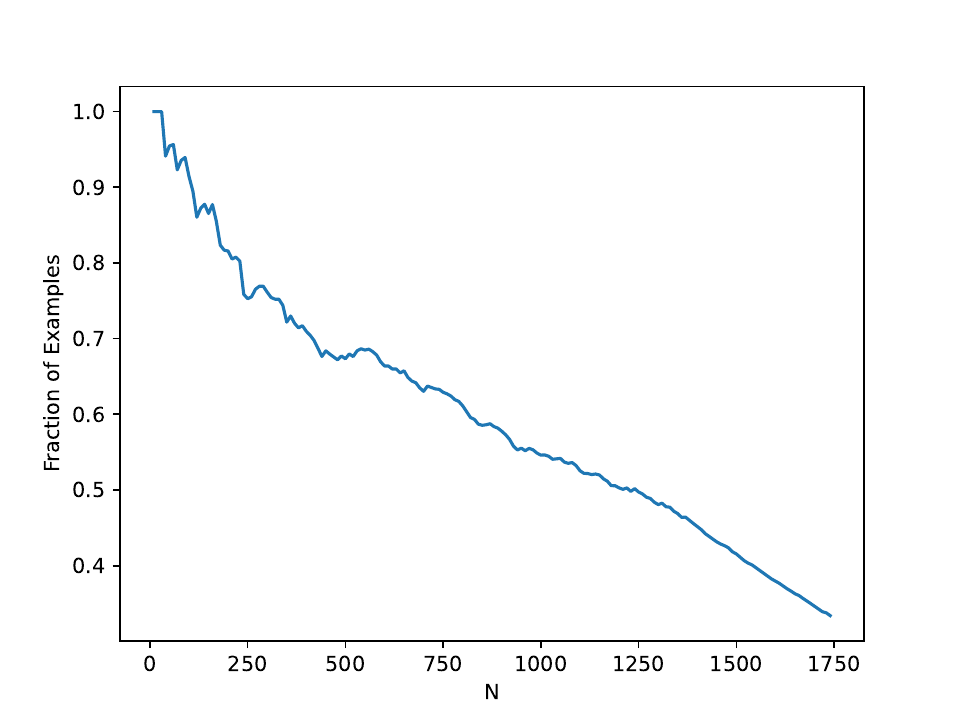}
    \caption{Fraction of error instances containing the word \textit{Cuba} in Top-N instances using $DM_mean$ score metric.}
    \label{fig:full_cuban_word_error}
\end{figure}

\begin{figure}[htb!]
    \centering\includegraphics[width=\columnwidth]{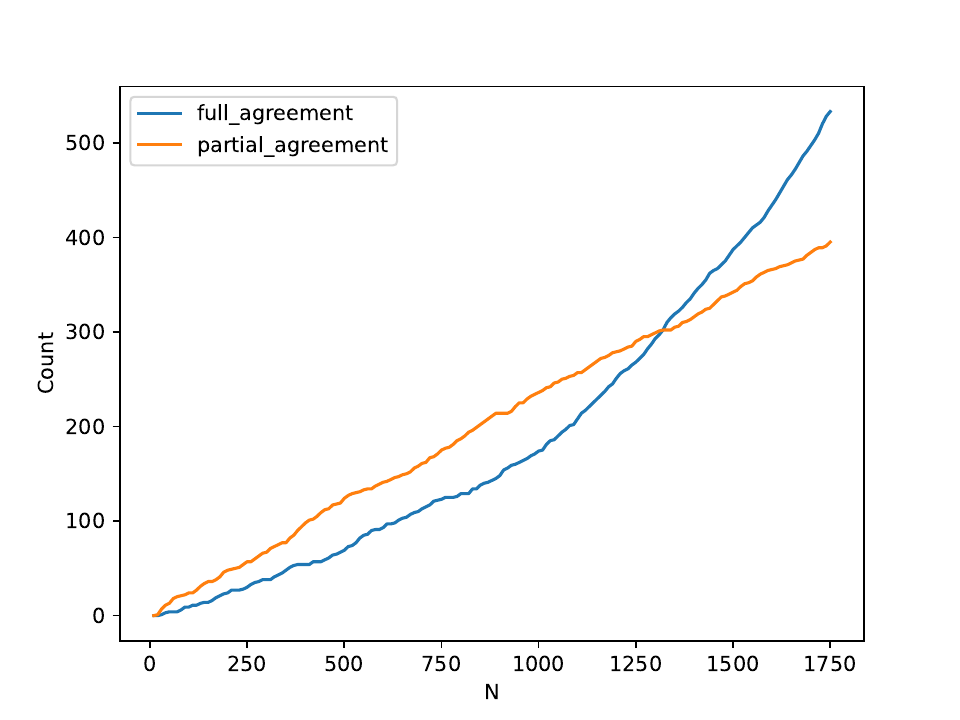}
    \caption{Agreement index for error instances in Top-N using $DM_{mean}$ score metric.}
    \label{fig:agreement_errors}
\end{figure}

In the \cupansp dataset, Figure \ref{fig:full_cuban_word_error} shows that about 67\% of the Top-500 non-common examples and 54\% of the Top-1000 non-common examples contained the word \textit{Cuba}, suggesting a strong influence on model behavior, given that only 33\% of the total examples contain this word.
Additionally, we found that 63.31\% of the Top-500 errors in \cupansp were cases where only two annotators agreed on the label, and for the Top-1000, this number was 57\%. Across all non-common instances, full agreement (three annotators) occurred in 57\% of cases, indicating a clear link between annotation difficulty and model errors as shown in Figure \ref{fig:agreement_errors}.
\begin{figure*}[htb!]
    \centering
    \begin{subfigure}[t]{\textwidth}  
         \centering
         \includegraphics[width=\textwidth]{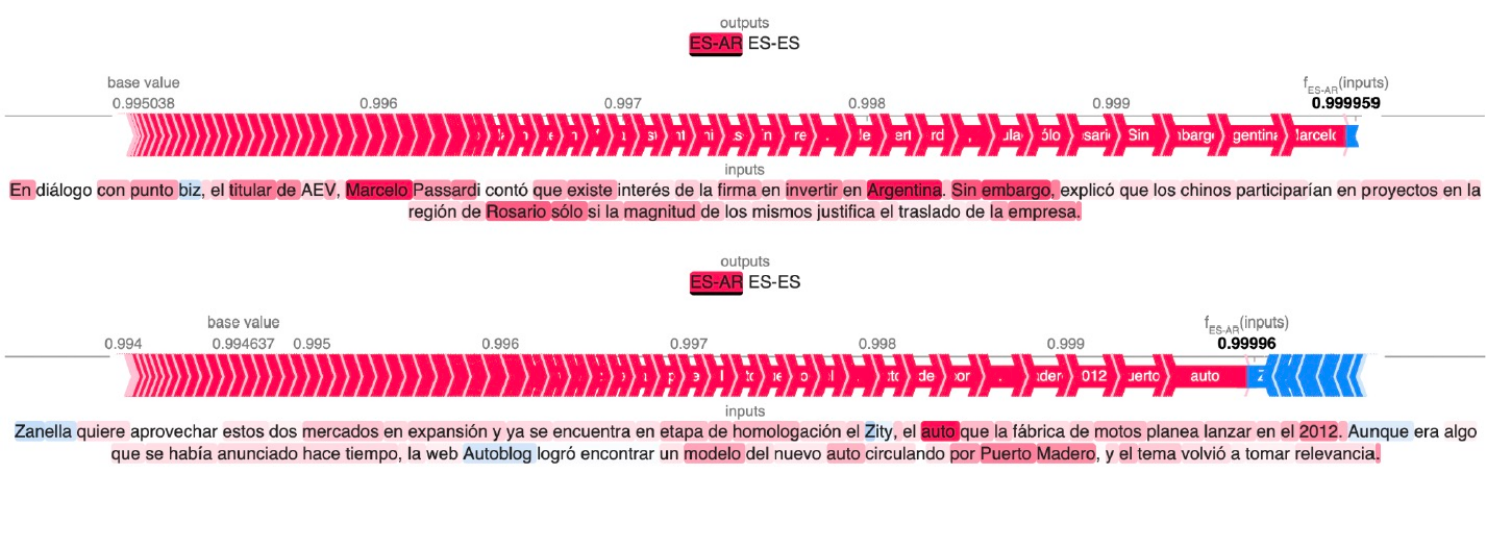}
         \caption{\dsltl dataset examples.}
         \label{fig:shap_dsl-tl}
    \end{subfigure}
    \hfill  
    \begin{subfigure}[t]{\textwidth}
         \centering
         \includegraphics[width=\textwidth]{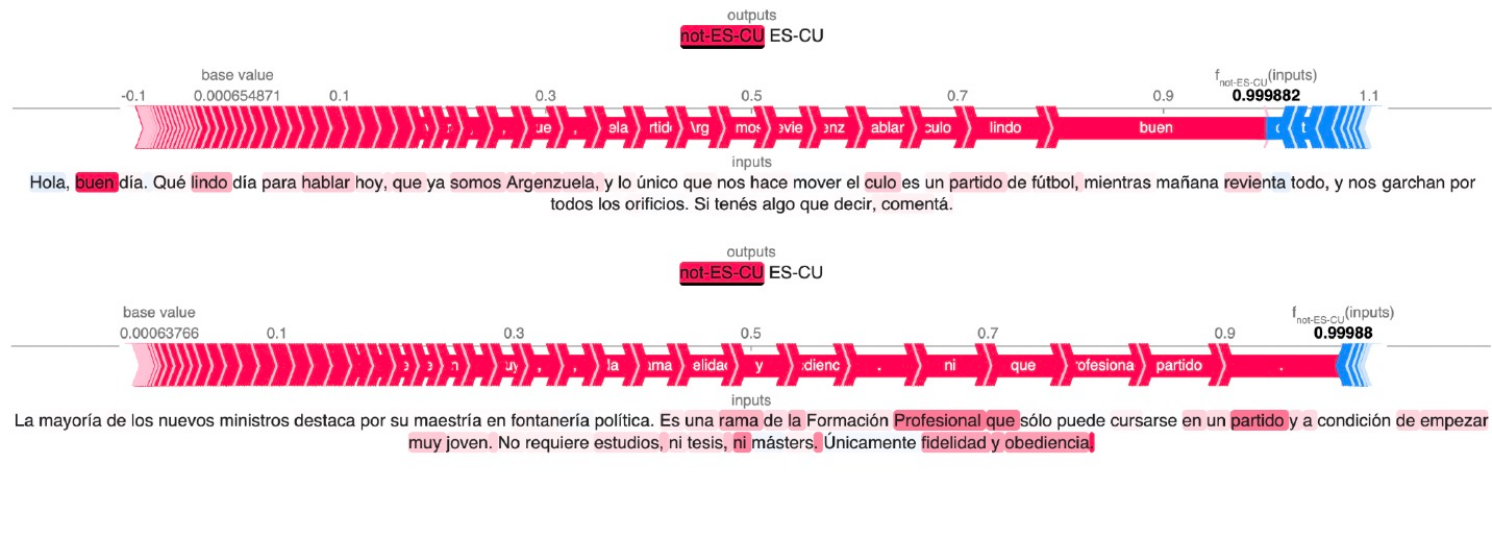}
         \caption{\cupansp dataset examples.}
         \label{fig:shap_cuban}
    \end{subfigure}
    \caption{For each dataset, we analyze the last two common examples in the ranking obtained using $DM_{mean-pred}$. The model is trained on binary classification for variety detection. The final output of the models for the predicted variety/class is highlighted. \textcolor{red}{Red}-colored terms influence the final decision towards \textit{ES-ES} or \textit{ES-CU} depending on the dataset, while \textcolor{blue}{blue}-colored terms influence the model classification towards \textit{ES-AR} or \textit{not-ES-CU}  classes.}
    \label{fig:shap_analyisis}
\end{figure*}

Another key point is understanding why the model fails to retrieve certain common examples. We focus on the last two common examples in the ranking for each dataset,  using SHapley Additive exPlanations (SHAP) \citep{lundberg2017unified} to analyze the model's behavior. SHAP is based on Game Theory and assigns importance scores to features, showing how much each feature influences the model's prediction.
Figure \ref{fig:shap_dsl-tl} presents the SHAP scores for the last two common examples in the \dsltl dataset ranking. For the first example, the words \textit{Argentina}, \textit{Rosario}, and \textit{Marcelo} are the most influential for predicting the \textit{ES-AR} label. The first two refer to the country and one of its major cities, while \textit{Marcelo} is a common name in Argentina. For the second example, \textit{auto} (commonly used in Argentina to mean "car," as opposed to \textit{coche} in Spain) is the most significant feature, followed by \textit{Puerto} and \textit{Madero}, a well-known place in Argentina. While named entities influence the first example, the second example, with the word \textit{auto}, suggests a potential annotation error, as it points to the Argentinian variety.

In Figure \ref{fig:shap_cuban}, we provide the corresponding analysis for the \cupansp dataset. For the first example, the word \textit{buen} (from the phrase \textit{buen día}, which is used in Spanish varieties other than the Cuban one) is the most significant, along with \textit{Argenzuela} (a blend of Argentina and Venezuela), \textit{garchan}, and \textit{tenés}, which are characteristic of the Argentinian variety. This example likely represents an annotation mistake. For the second example, the most influential words are \textit{profesional}, \textit{partido}, \textit{fidelidad}, and \textit{obediencia}, none of which are strong indicators of a specific variety. This suggests that common topics in Cuban tweets may affect the model's prediction, potentially introducing biases into the classification process.

Regarding the named entities, we followed the precedent set by previous works in variety Identification, such as the study introducing the \dsltl dataset \citep{zampieri-etal-2024-language}, retained named entities. Consequently, we included them in our initial approach \textbf{while emphasizing the importance of analyzing their influence}. We agree that a set of experiments where we could switch the named entities with neutral entities (or even adversarial entities (eg switch SosCuba with SosMexico) would be interesting. In our case, while evidence suggests that named entities contribute to model errors, \textbf{our preliminary analysis demonstrates the model's robustness to their presence}. For example, the sentence ``Mi mensaje para el pueblo de cuba emoji bandera cuba emoji :. ¡No están solos!. Cuenten con nosotros para seguir apoyando su lucha por la libertad y la democracia. soscuba url'' was ranked second using the Datamaps mean approach. Although it contained clear markers such as ``Cuba'' and ``soscuba,'' the model correctly identified it. This is not an isolated case, and further analysis of correctly classified examples can provide additional evidence of the system's robustness.

\section{Conclusion}
In this work, we examine the effectiveness of Datamaps methods in identifying common examples across closely related language varieties. Our results demonstrate the value of training dynamics in detecting difficult examples early in the model's learning process, as reflected by the effectiveness of $DM_{\text{mean-pred}}$ across both datasets. This confidence-based approach consistently outperformed the variability-based method, suggesting that tracking model confidence over predicted labels offers a reliable way to identify common examples automatically across different domains. Although the performance difference between variability-based and confidence-based approaches is less significant for theinformal dataset, the overall results indicate that confidence-based Datamaps can be a powerful tool for improving data quality in different contexts.

Although these methods may not fully solve the challenges of variety and dialect annotation, they offer a promising step forward, particularly when combined with automatic techniques and targeted human annotation.

\iffalse
We hope that this first dataset, freely available under a CC-BY-SA license upon publication, which focuses on Cuban, a Caribbean Spanish variety,  will prove a valuable resource for future research on that topic. 
\else

We hope that this initial dataset, freely accessible under a CC-BY-SA license upon publication, the first centered on Cuban, a Caribbean variety of Spanish, will prove  a valuable resource for future research on this topic.

\fi

\section{Limitations}

One limitation of our work is that the analysis focuses on binary classification scenarios, explicitly distinguishing between two main classes in each dataset without incorporating multi-class approaches or more complex variety distinctions. While this setup allows us to study common examples effectively, expanding the approach to multi-variety settings could provide a more comprehensive understanding of the challenges posed by language variety identification.

Another limitation is inherent in the way the annotations in the \cupansp dataset were built. Since all annotators were Cuban native speakers, the dataset focuses on Cuban versus non-Cuban distinctions. Incorporating annotators from other Spanish-speaking regions would allow for broader variety distinctions and more nuanced annotations, which could reduce potential biases introduced by a single-region perspective. \textbf{However, the framework for annotations was designed with enough flexibility to make it extensible for further annotations in variants different from Cuban} with the final aim of creating a dataset which cover most of the Spanish varieties. In this scenario, common examples between specific varieties will be determined by overlapping between  annotation made by native speakers from each variant.

Finally, as discussed in Section \ref{sec:errors}, named entities, including hashtags, play a significant role in model behavior. Managing these entities, such as replacing them with special tokens, could be an effective way to reduce bias and improve generalization. This is especially important in tasks like language variety classification, where named entities might disproportionately influence predictions.

\section{Ethical Considerations}

This work involves using social media data, particularly from Twitter, which may contain sensitive or controversial content. Although we anonymize the data by replacing user mentions and URLs, the content could still involve personal opinions, political statements, or even hate speech, especially in datasets like the \cupansp dataset, which includes tweets related to politically sensitive events such as the July 11th protests in Cuba. Given the nature of the protests, some tweets may contain offensive content. We are aware of the potential privacy implications when working with such data, and we have adhered to Twitter’s data usage policy to ensure compliance with ethical standards. Researchers accessing this dataset should consider the ethical implications when using politically charged content or messages that might harm individuals or communities.

Furthermore, identifying language varieties, especially in socially and politically sensitive contexts, risks reinforcing stereotypes or biases associated with particular regions. In this work, we frame our approach as a technical solution for linguistic diversity and not as a tool for making any sociopolitical or cultural assumptions about the speakers of these varieties. However, we acknowledge that any automated system trained on real-world data is susceptible to unintended biases arising from imbalanced datasets or biased annotations. The annotations in the \cupansp dataset are from native Cuban speakers, and while this helps in identifying Cuban Spanish, it may introduce a regional bias.
\section*{Acknowledgments}
This work received funding from the European Union’s Horizon 2020 research and innovation program under grant agreement No. 101021607. The authors warmly thank the OPAL infrastructure from Université Côte d'Azur for providing resources and support.
\bibliography{anthology,custom}

\appendix
\section{Data Preprocessing:}
\label{sec:data_processing}
Following previous works \citep{pérez2022robertuito,castillo-lopez-etal-2023-analyzing}, we pre-processed the data by replacing user mentions with the token \textit{@usuario} (or \textit{@user} in English), allowing up to two consecutive mentions. URLs were substituted with the token \textit{url}, and hashtags were segmented into words assuming Camel Case typing (e.g., \#CubaIslaBella becomes \textit{Cuba isla bella}). Emojis were replaced with their corresponding descriptions using the \textit{emoji} python library \footnote{\href{https://pypi.org/project/emoji/}{Emoji python library website.}}, and any repeated emojis were removed. Laughs were normalized to \textit{jaja}, following the standard in Spanish, and for letter repetitions, we kept up to two. We also removed repeated spaces and replaced line breaks with periods.

\begin{table}[h!]
\centering
\footnotesize
\begin{tabular}{ll}
\toprule
\#sentences & 1762 \\
\#tokens & 41374  \\
Avg length & 23.48 \\
Length variation (std) & 13.49 \\
Vocab size (unique words) &  13336  \\ 
\bottomrule
\end{tabular}
  \caption{\dsltl Overview.}
  \label{tab:dataset_overview}
\end{table}
\section{Annotation Guidelines for \cupansp}
\label{sec:annotation_guidelines}

The following guidelines were provided to the annotators to ensure consistent labeling of the dataset:

\begin{itemize}
    \item \textbf{cuban\_variety}: A boolean value indicating whether the tweet belongs to the target Spanish variety (Cuban). This value should be set to \textit{true} only if the annotator can clearly identify evidence that the tweet belongs to the Cuban variety.
    
    \item \textbf{not\_cuban\_variety}: A boolean value indicating that the tweet does not belong to the target Cuban variety. This value should be set to \textit{true} only if it is clear that the tweet does not belong to the Cuban variety, even if the specific variety cannot be identified.
    
    \item \textbf{specific\_variety}: A string indicating the specific variety if the annotator can easily identify it. The value should remain empty if the specific variety cannot be identified. The possible varieties are based on the Spanish varieties map presented in the appendix of \textit{Analyzing Zero-Shot Transfer Scenarios Across Spanish Variants for Hate Speech Detection}. These are:
    \begin{itemize}
        \item Other Caribbean variety
        \item Central American varieties (Costa Rica, El Salvador, Panamá)
        \item Mexican
        \item Spain
        \item Rioplatense (Argentina, Uruguay)
        \item Chilean
        \item Habla de las tierras altas (Perú, Venezuela, Colombia, Bolivia, Ecuador)
    \end{itemize}
    
    \item \textbf{unable\_to\_identify\_variety}: A boolean value set to \textit{true} if the annotator cannot identify any specific variety for the tweet.
    
    \item \textbf{irrelevant}: A boolean value set to \textit{true} if the tweet's content is considered irrelevant. This can be due to the tweet's size or other characteristics that lead to a lack of meaningful content.
\end{itemize}
\begin{table}[h!]
\centering
\begin{tabular}{ccc}
\hline
\textbf{Annotator} & \textbf{Age} & \textbf{Gender} \\ \hline
Annotator 1        & 26           & Male            \\ 
Annotator 2        & 26           & Female          \\ 
Annotator 3        & 23           & Female          \\ \hline
\end{tabular}
\caption{Socio-demographic attributes of the annotators}
\label{tab:annotators}
\end{table}

These annotations guidelines are extensible for speakers form varieties different from Cuba by changing the variety target. It makes it possible to extend the varieties covered in the dataset in a direct way.

\section{Hyper-parameters}

\begin{table}[h]
\centering
\begin{tabular}{ll}
\toprule
\textbf{Hyper-parameter}      & \textbf{Value}          \\
\midrule
Max sequence length           & 512                     \\
Batch size                    & 32                      \\
FP16                          & Enabled                 \\
Learning rate                 & 1e-5           \\
Epochs                        & 10                      \\
Scheduler                     & linear                  \\
Warmup ratio                  & 0.1                     \\
Weight decay                  & 0.01                    \\
Save strategy                 & Epoch                   \\
Logging steps                 & 10                      \\
Seed                          & \{42,151,2021,15,98\}         \\
\bottomrule
\end{tabular}
\caption{Hyper-parameters used for the fine-tuning.}
\label{tab:training_params}
\end{table}
The model will be released under the Creative Commons CC-BY-SA license, allowing for open access and use with appropriate attribution.

All experiments were conducted using a single NVIDIA RTX 8000 GPU, with each experiment taking less than two hours to complete. We used the \texttt{AutoModelForSequenceClassification} from Hugging Face's Transformers library \cite{wolf-etal-2020-transformers} for sequence classification tasks.

\section{Variety Identification Results}
\subsection{Variety Identification Benchmarks on \cupansp dataset}
In this section, we present the benchmark results for the \cupansp dataset. We use the same experimental setting for this task, as explained before. We present the dataset's benchmark for both approaches, single and multi-class. For the multi-class approach, we follow the procedure suggested by \citet{keleg-magdy-2023-arabic, bernier-colborne-etal-2023-dialect} of using one binary classifier per label. For the metrics, we used the macro average across all possible varieties.

Table \ref{table:benchmarks} shows the final results. We can notice a significant improvement in the model's performance in the multi-class scenario. This strengthens the point about single-class approach limitations for variety identification.

\begin{table*}[htb!]
\centering
\footnotesize
\begin{tabular}{ccccc}
\toprule

\multicolumn{1}{r}{Approach} & Acc & Precision & Recall & f1-score \\ \hline
single-class                        & 67.54 ± 1.42 & 65.86 ± 1.69 & 64.45 ± 1.01 & 64.62 ± 1.05 \\

multi-class                          & \textbf{78.69 ± 0.86} & \textbf{82.64 ± 0.91} & \textbf{87.80 ± 1.28} & \textbf{85.06 ± 0.61} \\
\bottomrule
\end{tabular}
\caption{Benchmarks for Variety Identification task on \cupansp dataset. We present the results for both the single-class and the multi-class approaches. }
\label{table:benchmarks}
\end{table*}

\subsection{Variety Identification Benchmarks on \dsltl dataset}

In this section, we present the benchmark results for the \dsltl dataset. Table \ref{table:benchmarks_dsl} shows the final results. As for the \cupansp dataset, there is a significant improvement in the model's performance in the multi-class scenario.

\begin{table*}[htb!]
\centering
\footnotesize
\begin{tabular}{ccccc}
\toprule

\multicolumn{1}{r}{Approach} & Acc & Precision & Recall & f1-score \\ \hline
single-class                        & 76.76 ± 0.74 & 76.18 ± 0.80 & 75.78 ± 0.75 & 76.76 ± 0.74 \\

multi-class                          & \textbf{77.65 ± 0.27} & \textbf{82.00 ± 0.29} & \textbf{83.99 ± 0.30} & \textbf{82.97 ± 0.25} \\
\bottomrule
\end{tabular}
\caption{Benchmarks for Variety Identification task on \dsltl dataset. We present the results for both the single-class and the multi-class approaches. }
\label{table:benchmarks_dsl}
\end{table*}

\end{document}